\documentclass[10pt,twocolumn,letterpaper]{article}

\usepackage{cvpr}              

\usepackage{multirow}
\usepackage{amsmath}
\usepackage{graphicx} 
\usepackage{booktabs} 
\usepackage{bbding}
\usepackage{colortbl}

\definecolor{cvprblue}{rgb}{0.21,0.49,0.74}
\usepackage[pagebackref,breaklinks,colorlinks,allcolors=cvprblue]{hyperref}

\title{MMTL-UniAD: A Unified Framework for Multimodal and Multi-Task Learning in Assistive Driving Perception}

\author{%
Wenzhuo Liu$^{1}$~~~~~ Wenshuo Wang\footnotemark[1]~$^{,1}$~~~~~ Yicheng Qiao$^2$~~~~~  Qiannan Guo$^{2}$~~~~~ Jiayin Zhu$^{3}$ ~~~~~ Pengfei Li$^{2}$ \\ \vspace{0.5em}Zilong Chen$^{2}$~~~~~ Huiming Yang$^{2}$~~~~~ Zhiwei Li$^{4}$~~~~~  Lening Wang$^{5}$~~~~~ Tiao Tan$^{2}$~~~~~ Huaping Liu$^{2}$\\ 
$^1$Beijing Institute of Technology, Zhuhai~~~~~~ $^2$Tsinghua University~~~~~~ $^3$HKUST(GZ)  \\ $^4$ Beijing University of Chemical Technology~~~~~~ $^5$Beihang University
}

\begin{document}
\maketitle

\renewcommand{\thefootnote}{\fnsymbol{footnote}}
\footnotetext[1]{Corresponding author.}

\begin{abstract}
Advanced driver assistance systems require a comprehensive understanding of the driver's mental/physical state and traffic context but existing works often neglect the potential benefits of joint learning between these tasks. This paper proposes MMTL-UniAD, a unified multi-modal multi-task learning framework that simultaneously recognizes driver behavior (e.g., looking around, talking), driver emotion (e.g., anxiety, happiness), vehicle behavior (e.g., parking, turning), and traffic context (e.g., traffic jam, traffic smooth). A key challenge is avoiding negative transfer between tasks, which can impair learning performance. To address this, we introduce two key components into the framework: one is the multi-axis region attention network to extract global context-sensitive features, and the other is the dual-branch multimodal embedding to learn multimodal embeddings from both task-shared and task-specific features. The former uses a multi-attention mechanism to extract task-relevant features, mitigating negative transfer caused by task-unrelated features. The latter employs a dual-branch structure to adaptively adjust task-shared and task-specific parameters, enhancing cross-task knowledge transfer while reducing task conflicts. We assess MMTL-UniAD on the AIDE dataset, using a series of ablation studies, and show that it outperforms state-of-the-art methods across all four tasks. The code is available on \url{https://github.com/Wenzhuo-Liu/MMTL-UniAD}.
\end{abstract}    
\section{Introduction}
\label{sec:intro}

\begin{figure}
\centering
\includegraphics[width=0.49\textwidth]{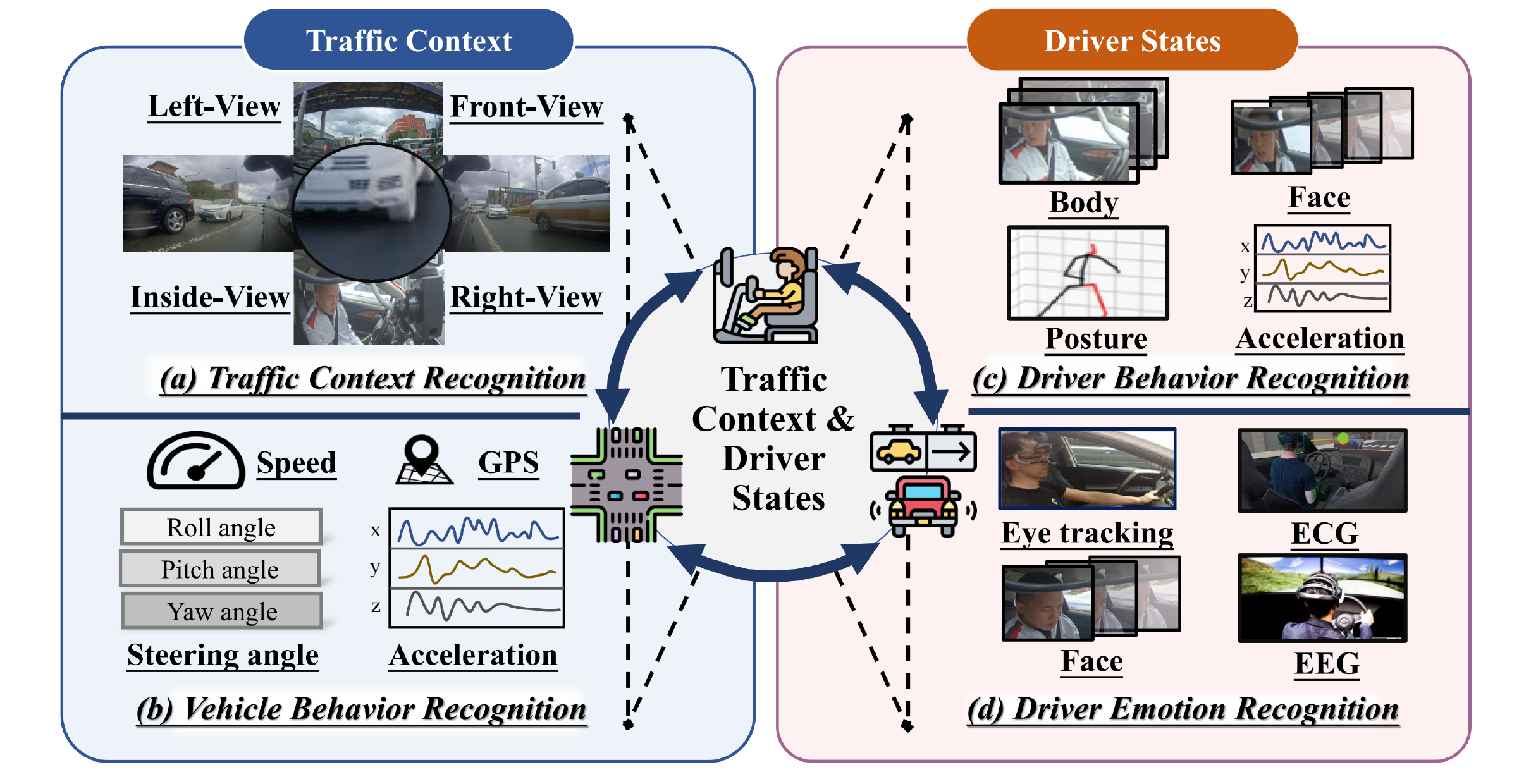}
\vspace{-1em}
\caption{Traffic context and driver states interaction diagram. Tasks (a), (b), (c), and (d) represent traffic context recognition, vehicle behavior recognition, driver behavior recognition, and driver emotion recognition, respectively. These tasks comprehensively demonstrate the complex and closely interconnected relationships between the driver and traffic.}
\label{network1}
\vspace{-1em}
\end{figure}

Over the past decade, Advanced Driver Assistance Systems (ADAS) --- such as automatic emergency braking, lane keeping assist, and surround view monitoring --- have significantly enhanced driving safety through the advances in monitoring driver states and surrounding traffic \cite{wang2023openoccupancy,zhang2023oblique,li2024mipdmultisensoryinteractiveperception,chen2024fw}. Despite these advancements, approximately 1.35 million fatalities occur annually in traffic accidents \cite{world2019global}, with human drivers' abnormal mental or physical states contributing to over 65\% of these incidents \cite{tian2013studying}. Consequently, accurate identification of driver states is essential for ADAS \cite{martin2019drive,li2021spontaneous} but remains a complex challenge due to the intricate causal relationships between driver states and traffic context \cite{yang2023aide} (see Fig. \ref{network1}). For instance, traffic congestion can induce driver anxiety, effecting driving behavior \cite{yang2023aide}.

Most existing approaches to driver state and traffic context recognition focus on isolated tasks, such as driver behavior or emotion recognition \cite{saleh2017driving,gong2023sifdrivenet,li2021cogemonet,zepf2020driver,yang2023robust,mou2023driver}, traffic environment recognition \cite{guo2023temporal,zhi2021bigru,martin2018dynamics,wang2023path}. However, they fail to exploit the interaction between these tasks, which can limit the potential for cross-task learning \cite{qian2019dlt}. In reality, these tasks are not independent when driving in real-world traffic \cite{yang2023aide,martin2019drive}. For example, lane-change behavior relies on the interaction between the current traffic context (e.g., congestion) and the driver's emotional state \cite{guo2023temporal,xing2021multi}.

Multi-task learning (MTL) offers an opportunity to enhance task recognition accuracy by enabling information sharing between related tasks \cite{chowdhuri2019multinet,ishihara2021multi}. However, current research is often hindered by negative transfer, a phenomenon where performance degrades due to information sharing between weakly related tasks. Many existing works focus on tasks with fewer differences. For instance, studies often combine similar tasks like lane detection, depth estimation, and drivable area segmentation \cite{wu2022yolop,qian2019dlt,teichmann2018multinet} to improve traffic context understanding, while others focus on joint learning for driver-related tasks such as behavior, emotion, intention recognition \cite{xing2021multi,kim2020multi,xun2020multitask}. However, neglecting the interaction between driver-related tasks and traffic context recognition limits the integration of driver state and environmental information, thus restricting ADAS's ability to fully understand driving context \cite{yang2023aide,martin2019drive}.

Another significant limitation of current MTL models is their reliance on single-modal inputs. Achieving satisfactory recognition performance typically requires the complementarity of multimodal data \cite{li2021cogemonet,du2020convolution,guo2023temporal,liu2024glmdrivenet,liu2024fmdnet,cui2024textnerf,gong2022multi}. For instance, traffic context recognition often relies on multi-view driving scene images, while driver behavior and state recognition necessitate comprehensive in-vehicle images, fine-grained driver images, and joint data for driver posture and gesture. Although leveraging these multimodal inputs can enhance the accuracy \cite{rong2020driver,guo2023temporal}, most existing MTL models rely on a single input type (e.g., driving scene images or driver images) \cite{choi2024multi,qian2019dlt,teichmann2018multinet}, severely limiting their practical application in ADAS.

To address these challenges, we propose a Unified Framework for Multimodal and Multi-Task Learning in Assistive Driving Perception (MMTL-UniAD). Our framework leverages multimodal data to simultaneously recognize driver behavior, emotion, traffic context, and vehicle behavior. First, inspired by \cite{wang2020axial,zhu2023biformer}, we designed a multi-axis region attention network for multi-view images from the driving environment and the human driver. This network captures global context through horizontal-vertical attention and then uses region attention to extract interest-triggering features, selecting task-related high-level semantic information. This approach mitigates negative transfer between tasks. Additionally, we introduce a dual-branch multimodal embedding that uses soft parameter sharing \cite{ma2018modeling} to adaptively adjust task-shared and task-specific parameters. This design enhances task-specific learning while promoting information sharing and positive transfer between tasks. We validate our approach on the publicly available AIDE dataset, demonstrating superior performance over state-of-the-art methods. Our contributions are:

\begin{itemize}
    \item [$\bullet$] We propose MMTL-UniAD, a novel framework that addresses the challenges of multimodal multi-task learning (MMTL) in assistive driving.
    \item [$\bullet$] We designed a multi-axis region attention network to effectively extract features from multi-view images of the driving environment and human drivers.
    \item [$\bullet$] We introduce a dual-branch multimodal embedding to extract task-shared and task-specific features, mitigating negative transfer caused by task differences while enhancing cross-task knowledge learning.
\end{itemize}
\section{Related Work}
\label{sec:Related Work}

\subsection{Multi-task Learning}

Recent advancements in deep neural networks have made it feasible to learn multiple tasks jointly, improving the performance of individual tasks through shared learning. A key strategy for achieving multi-task learning is the sharing of parameters across tasks. Two primary approaches for parameter sharing are hard parameter sharing and soft parameter sharing. 
\begin{itemize}
    \item Hard parameter sharing limits the sharing of parameters to the \textit{output} layer of the neural network \cite{li2020knowledge,xu2018pad,cao2023relational,al2022zero,ghiasi2021multi,gan2024segmentation,huang2024mfe}. This method effectively reduces overfitting between tasks and improve learning efficiency. However, it lacks flexibility, as it heavily depends on the correlations between tasks. If the task has significantly different objectives or conflicts, the shared parameters may harm the performance of individual tasks \cite{misra2016cross}.
    \item Soft parameter sharing, on the other hand, extends sharing to both \textit{output} and \textit{internal} layers of the network \cite{gao2023enhanced,chen2023adamv,xu2023demt,xu2022mtformer}. This approach helps reduce conflicts between tasks, even when the tasks have weak correlations. By allowing more flexible adaptation, soft parameter sharing can enhance the performance and adaptability of each individual task, particularly in settings where tasks may differ in complexity or relevance.
\end{itemize}
In the context of MTL for ADAS, soft parameter sharing has become an attractive choice for improving the overall performance across multiple driving-related tasks by allowing tasks to learn shared representations while maintaining flexibility in task-specific learning.

\subsection{Driver State Recognition}

Driver state recognition is a critical component of ADAS, as understanding the mental and physical state of the driver is essential for ensuring road safety. Existing studies have leveraged various signals to infer driver states, focusing on both driver-specific and traffic-related information. 
\begin{itemize}
    \item Some approaches rely on vehicle dynamics-based data (e.g., speed, steering angle) alongside multi-view images (e.g., front-view, right-view, and left-view) of the driving scene to assess the driver's state \cite{saleh2017driving,gong2023sifdrivenet,liu2024glmdrivenet,liu2024fmdnet}. These methods focus on the relationship between the vehicle's movement and the surrounding environment but often fail to capture the driver's internal state. 
    \item Other studies have utilized driver behavior data, including images of the driver and physiological signals (e.g., heart rate, eye gaze) to detect emotional or cognitive states such as fatigue, stress, or distraction. \cite{li2021cogemonet,du2020convolution,zepf2020driver,yang2023robust,mou2023driver}. These approaches focus more directly on the driver but may neglect the impact of the surrounding traffic context on driver behavior.
\end{itemize}
\begin{figure}[t]
    \centering
    \includegraphics[width=0.49\textwidth]{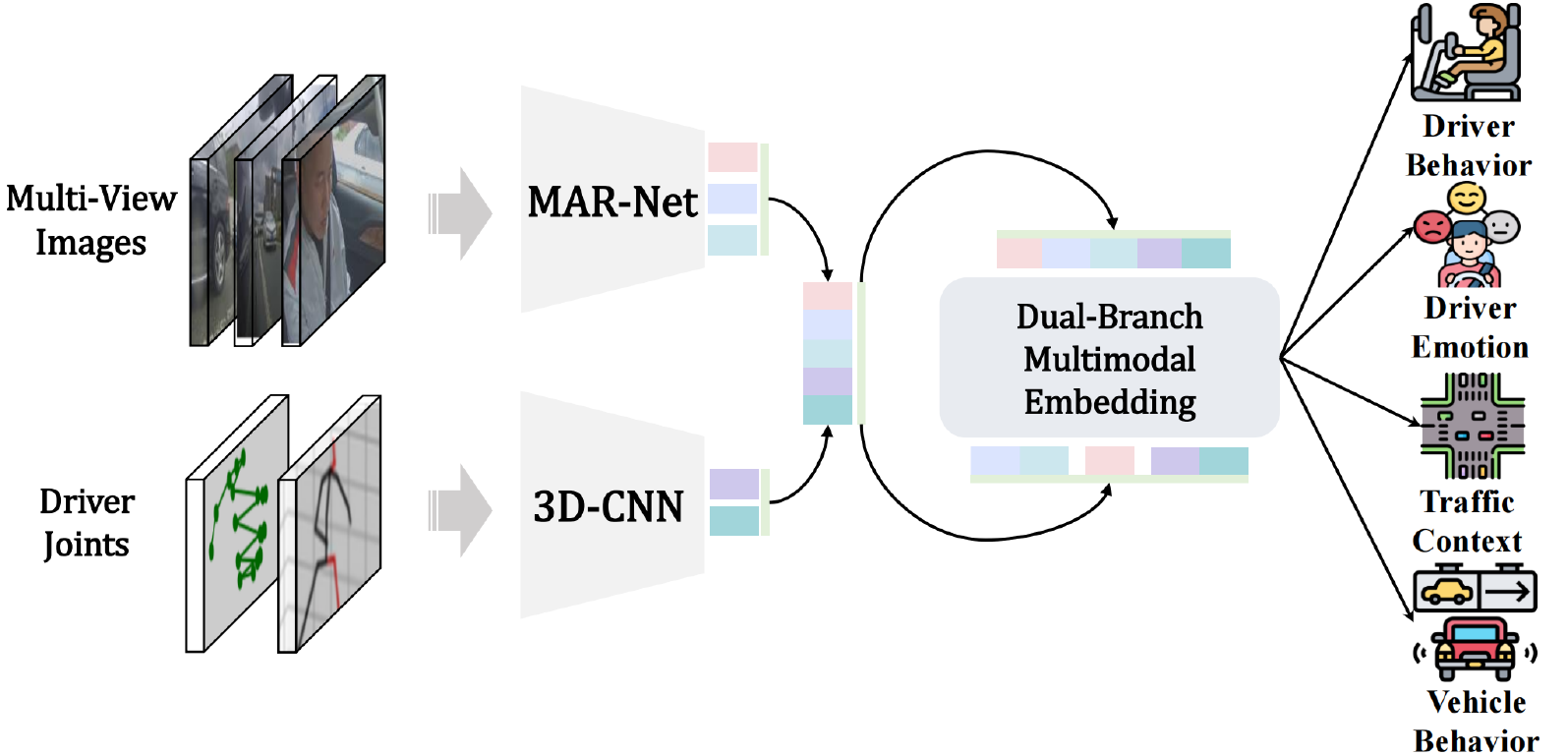}
    \vspace{-1em}
    \caption{The overall pipeline of MMTL-UniAD. MMTL-UniAD consists of two primary components: Multimodal Encoder and Dual-Branch Multimodal Embedding. The multimodal encoder is composed of a Multi-axis Regional Attention Network (MARNet) and a 3D-CNN, which are responsible for extracting features from multi-view images and driver joint, respectively. The Dual-Branch Multimodal Embeddings further integrate the multimodal features for multi-task recognition.}
    \label{fig:network-overview}
\end{figure}
While these works have made progress in recognizing driver states, they tend to focus either on driver information or traffic context, without effectively combining both. This limitation becomes apparent in complex, dynamic driving situations, where both the driver's internal state and external traffic context jointly influence driving behavior. As such, integrating driver states and traffic context remains an open challenge in ADAS research, as it requires the simultaneous consideration of both domains to provide a comprehensive understanding of the driving context.

\section{Methods}

This section presents the proposed MMTL-UniAD framework (Fig. \ref{fig:network-overview}), highlighting its two key components: Multi-axis Region Attention Network (MARNet) and Dual-Branch Multimodal Embedding.

\subsection{Network Overview}

The framework of MMTL-UniAD (Fig. \ref{fig:network-overview}) consists of two main modules: the multimodal encoder and dual-branch multimodal embedding. The former is composed of a Multi-axis Region Attention Network (MARNet) and a 3D Convolutional Neural Network (3D-CNN). Specifically, MARNet captures key features from multi-view images (i.e., front-view, right-view, left-view, inside-view, driver face, and driver body) through multi-attention mechanisms, while the 3D-CNN extracts prominent features from driver joint data (i.e., gesture and posture). The latter consists of a task-shared branch and a task-specific branch, which are used to further fuse the extracted multimodal features. This module first extracts task-shared features $\mathbf{F}_{\mathrm{sh}}$ and task-specific features $\mathbf{F}_{\mathrm{sp}}$ by adaptively adjusting the parameters of these two branches, enhancing cross-task knowledge sharing while capturing unique information of tasks. Next, the two features are integrated through dynamic fusion, to obtain the recognition results $O_{j}$, for individual tasks $j$: driver behavior recognition, driver emotion recognition, traffic context recognition, and vehicle behavior recognition. The recognition process for each task is as follows:
\begin{equation}
\label{Oj}
O_j = \sigma(w_j) L^{1}_j(\mathbf{F}_{sh}) + (1 - \sigma(w_j)) L^{2}_j(w_{ca}(\mathbf{F}_{sp_j})),
\end{equation}
where $L^{1}_j$ and  $L^{2}_j$  are fully connected layers corresponding to task $j$, $w_{ca}$ represents a channel attention weighting operation that prioritizes relevant features for the task, $w_j$ is a learnable weight parameter that dynamically selects the most effective information from task-shared and task-specific features, and $\sigma$ denotes the Sigmoid function. This formulation enables the model to adaptively prioritize shared knowledge across tasks, while retaining the distinct characteristics necessary for task-specific performance.

\begin{figure}
\centering
\includegraphics[width=0.48\textwidth]{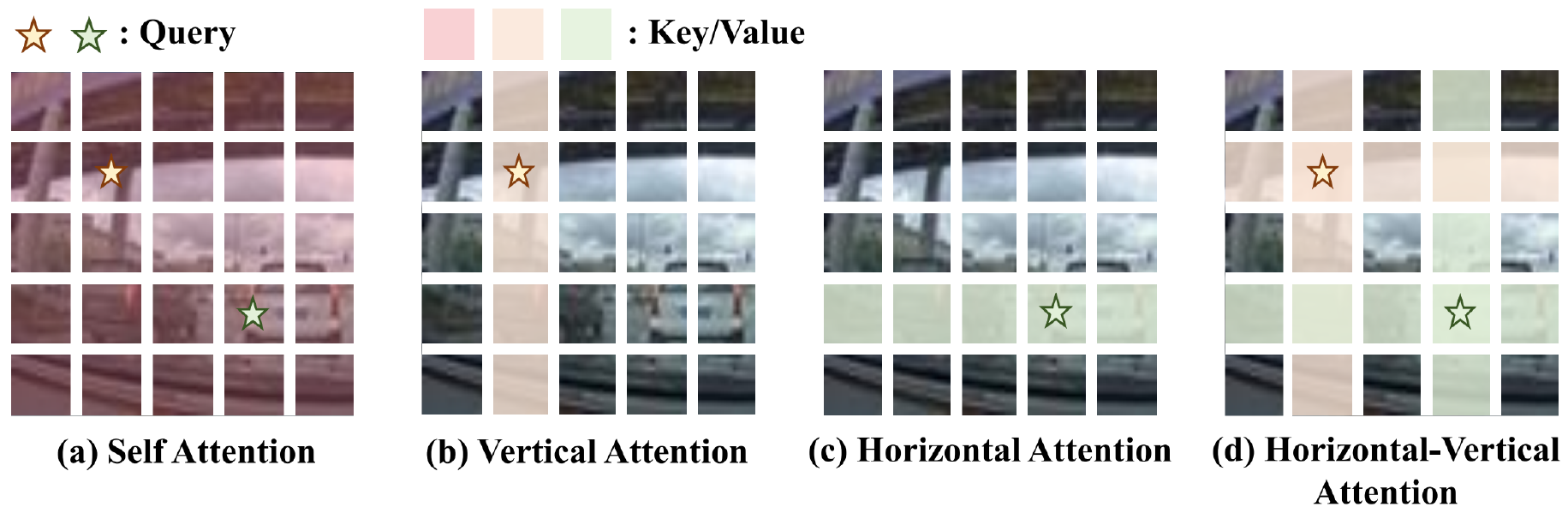}
\vspace{-1em}
\caption{Diagram of different self-attention. (a) represents the most common global self-attention in images; (b) (c) (d) representing vertical attention, horizontal attention and horizontal-vertical attention respectively. Among them (d) represents the horizontal-vertical attention we introduced.}
\vspace{-1em}
\label{fig5}
\end{figure}

\subsection{Multimodal Encoder}

\subsubsection{Multi-axis Region Attention Network}

\begin{figure*}[h]
    \centering
    \includegraphics[width=0.98\textwidth]{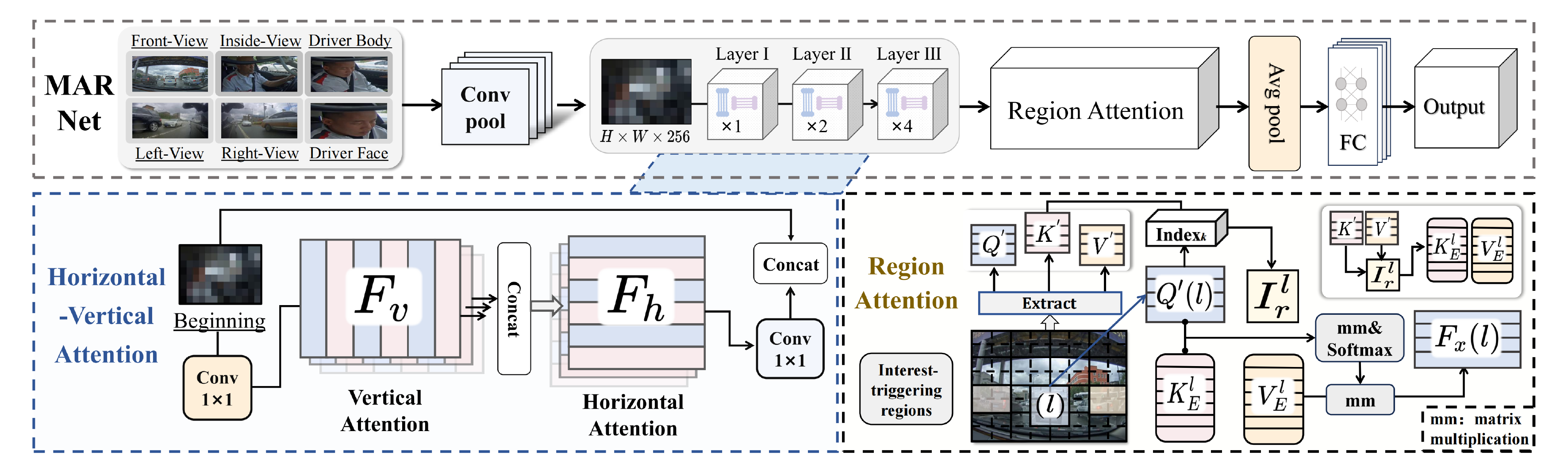}
    \caption{The flowchart of the MARNet architecture, including the processes for horizontal-vertical attention and region attention.}
    \label{fig:GCFANet}
    \vspace{-1em}
\end{figure*}

Multi-view images of driving environment and human drivers usually contain many task-unrelated features such as billboards along the roadside and in-car decorative items. In multi-task learning, the quality of extracted features directly impacts cross-task synergy. Selecting features relevant to tasks can facilitate information complementarity between tasks in feature sharing; otherwise, would lead to a negative transfer issue. To address this challenge, we designed MARNet, as shown in Fig. \ref{fig:GCFANet}. In this network, task-related features from multi-view images are extracted using the proposed horizontal-vertical attention and region attention to alleviate the negative transfer issue caused by irrelevant features across tasks. 

Let $\mathbf{F}_o \in \mathbb{R}^{H \times W \times C}$ be the input feature map, which is obtained from the multi-view images through initial convolution operations, and $H$, $W$ and $C$ are the height, width, and number of channels of the map. The horizontal-vertical attention first performs linear projections on the input feature map $\mathbf{F}_o$ using three weight matrices to generate the query, key, and value, denoted as $\mathbf{Q}$, $\mathbf{K}$, and $\mathbf{V}$, respectively. We then apply self-attention along the vertical direction of $\mathbf{F}_o$ at each position ($h,w$) to integrate features relevant (see Fig. \ref{fig5} (b)), resulting in a new feature $\mathbf{F}_v$, the computation process is as follows:
\begin{equation}
\begin{aligned}
\mathbf{F}_v &= \sum_{w=1}^{W} \sum_{h=1}^{H} \sum_{h'=1}^{H} \text{softmax}\left( \frac{ \mathbf{Q}_{(h, w, \cdot)}  \mathbf{K}_{(h', w, \cdot)}^\top }{ \sqrt{C} } \right) \mathbf{V}_{(h', w, \cdot)},
\end{aligned}
\end{equation}
where $\mathbf{Q}_{(h, w, \cdot)}$ denotes the query vector at position $(h, w)$, $\mathbf{K}_{(h', w, \cdot)}$ and $ \mathbf{V}_{(h', w, \cdot)}$ denote the key and value vector at position $(h', w)$, respectively. Similarly, we can obtain $\mathbf{F}_h$ along the horizontal direction (see Fig. \ref{fig5} (c)), the computation process is as follows:
\begin{equation}
\begin{aligned}
\mathbf{F}_h &= \sum_{h=1}^{H} \sum_{w=1}^{W}  \sum_{w'=1}^{W} \text{softmax}\left( \frac{ \mathbf{F}_{v_{(h, w, \cdot)}}  \mathbf{K}_{(h, w', \cdot)}^\top }{ \sqrt{C} } \right) \mathbf{V}_{(h, w', \cdot)}.
\end{aligned}
\end{equation}

Next, we concatenate features $\mathbf{F}_h$ and $\mathbf{F}_o$ to obtain a new feature map $\mathbf{F}’ \in \mathbb{R}^{H \times W \times C}$. 
\begin{equation}
\mathbf{F}' = \mathrm{Concat}(\mathrm{C2D_{1*1}}(\mathbf{F}_h), \mathbf{F}_o),
\end{equation} 
where $\mathrm{C2D_{1*1}}(\cdot)$ denotes a 2D convolution operation  with kernel size of 1, and $\mathrm{Concat}(\cdot)$ denotes a concatenate operation. This operation utilizes the long-range dependencies extracted along horizontal-vertical directions of $\mathbf{F}_o$ (see Fig. \ref{fig5} (d)), preserving the detailed information of initial features.

Horizontal-vertical attention captures directional features and global context in multi-view images. However, in real-world scenarios, nearby road users (e.g., other vehicles and pedestrians) often appear in varying orientations, making it challenging to capture their dependencies effectively \cite{zhu2023biformer}. To address this limitation, we integrate MARNet with region attention, which enables the model to focus on the most relevant regions of the input feature map, $\mathbf{F}'$. This approach dynamically selects  regions based on similarity measures, allowing the model to emphasize interest-triggering features of dynamic objects that do not adhere to a fixed direction. 

Specifically, we first use region attention to partition the feature map $\mathbf{F}'$ into $\frac{HW}{t^2}$ independent regions, each of size $t \times t$. This transforms the feature map into $\mathbf{F}'' \in \mathbb{R}^{\frac{HW}{t^2} \times t^2 \times C}$. We then apply three weight matrices to linearly project $\mathbf{F}''$ into the query, key, and value, denoted as $\mathbf{Q}'$, $\mathbf{K}'$, $\mathbf{V}' \in \mathbb{R}^{\frac{HW}{t^2} \times t^2 \times C}$, respectively. To improve computational efficiency, we apply global average pooling across the second dimension of both $\mathbf{Q}'$ and $\mathbf{K}'$, resulting in $\mathbf{Q}''$, $\mathbf{K}'' \in \mathbb{R}^{\frac{HW}{t^2} \times C}$. The similarity matrix is then computed using the dot product between $\mathbf{Q}''$ and $\mathbf{K}''$, and the $k$ most similar regions for each region $l$ are selected, forming the index set $\mathbf{I}_r^l \in \mathbb{R}^{k}$:
\begin{equation}
\mathbf{I}_r^l = \text{Index}_k\left( \mathbf{Q}''(l) \mathbf{K}''^\top \right ),
\end{equation}
where $\mathbf{Q}''(l) \in \mathbb{R}^{1 \times C}$ is the query vector for region $l$, and $\mathbf{K}''^\top \in \mathbb{R}^{C \times \frac{HW}{t^2}}$ is the transpose of $\mathbf{K}''$. Using this index, we extract the corresponding rows from $\mathbf{K}'$ and $\mathbf{V}'$ to form $\mathbf{K}_E^l \in \mathbb{R}^{k \times t^2 \times C}$ and $\mathbf{V} _E^l\in\mathbb{R}^{k \times t^2 \times C} $, respectively. Attention is then computed for each region $l$ and its top $k$ most similar regions:
\begin{equation}
\mathbf{F}_x(l) = \text{softmax}\left( \frac{ \mathbf{Q}'(l) (\mathbf{K}_E^l)^\top }{ \sqrt{C} } \right) \mathbf{V}_E^l,
\end{equation}
where $\mathbf{F}_x(l) \in \mathbb{R}^{t^2 \times C}$ is the updated feature for region $l$ after weighting the features via attention. After this operation, the output feature for all regions is $\mathbf{F}_x \in \mathbb{R}^{\frac{HW}{t^2} \times t^2 \times C}$. Finally, we reshape  $\mathbf{F}_x$ back to the dimension $(H, W, C)$, apply global average pooling, and pass the pooled result through a fully connected layer to obtain the final feature representation $\mathbf{F}_i$, where $i \in \{1, 2, 3, 4, 5, 6\}$ corresponds to the six multi-view image inputs.

MARNet extracts deep features from multi-view images by combining horizontal-vertical attention with region attention, yielding a rich set of effective features for subsequent multimodal feature embedding.

\subsubsection{3D-CNN}

In the multimodal encoder, we use a 3D convolutional neural network (3D-CNN) to analyze the driver's gestures and postures from continuous video frames. The 3D-CNN captures spatiotemporal features by applying 3D convolutional kernels across temporal and spatial dimensions. The feature representation is denoted as $\mathbf{F}_i$, where $i \in \{7,8\}$ represents two joint inputs. The extracted features are primarily used to understand driver behavior and emotion.

\subsection{Dual-Branch Multimodal  Embedding}
\begin{figure}
\centering
\includegraphics[width=0.48\textwidth]{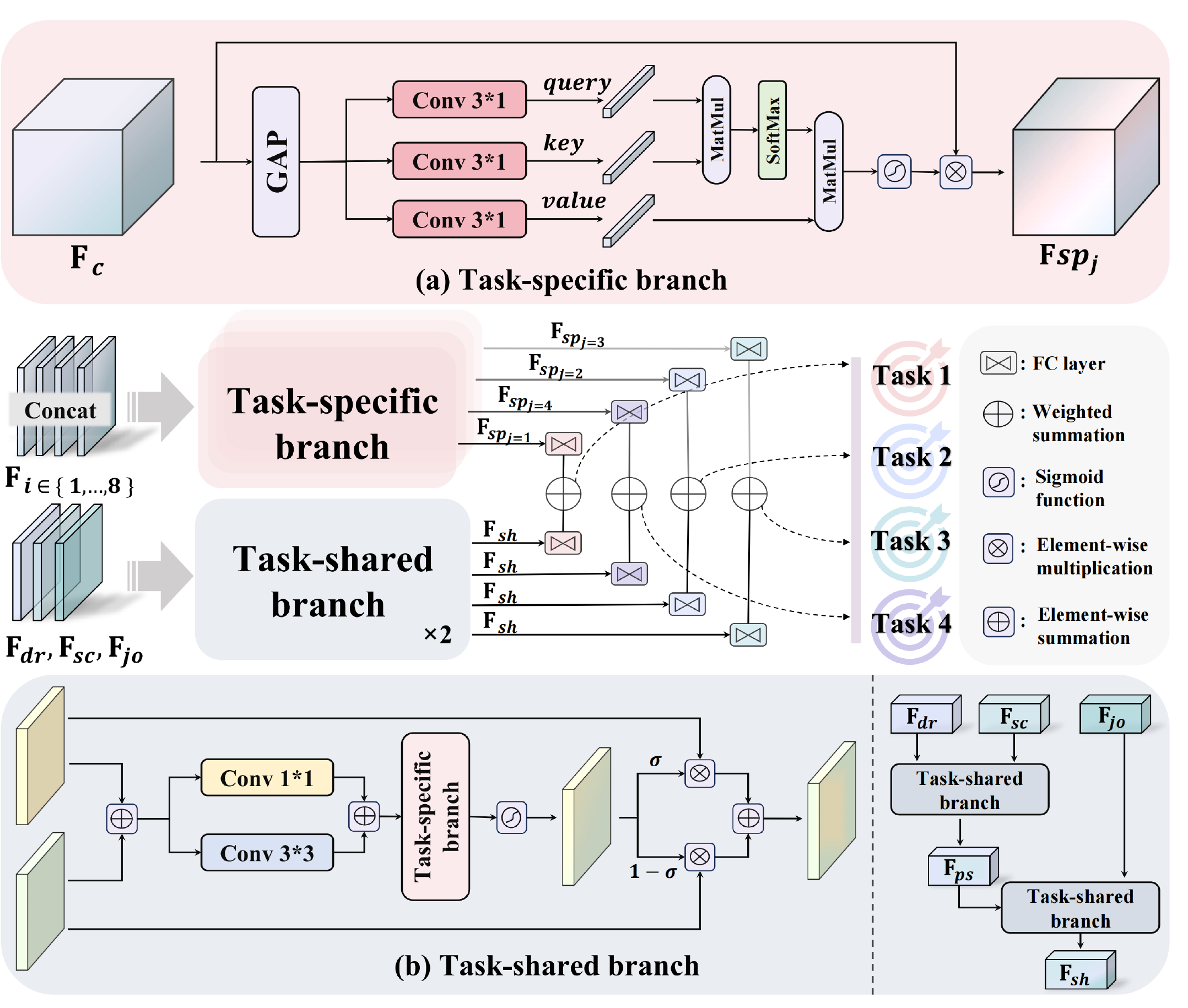}
\vspace{-1em}
\caption{Structural of Dual-Branch Multimodal Embedding.}
\vspace{-1em}
\label{SME}
\end{figure}

Balancing the synergy effect of task-shared and task-specific features is crucial in multi-task learning. Task-shared features enable knowledge transfer across tasks, promoting generalization \cite{cao2023relational,al2022zero}. However, disparities between tasks can result in negative transfer. Task-specific features help mitigate task conflict and reduce the risk of negative transfer. Yet, over-reliance on these features can limit information sharing, weakening the model's ability to generalize across tasks \cite{chen2023adamv,xu2023demt}. To address this, we designed a dual-branch multimodal embedding (Fig.\ref{SME}), which simultaneously extracts both feature types and adaptively balances their contributions based on the specific tasks at hand.

The dual-branch multimodal embedding consists of two primary components: the task-shared branch and the task-specific branch. To address the negative transfer due to inter-task differences, the task-specific branch extracts task-specific features from multimodal inputs, as shown in Fig.\ref{SME} (a). First, global average pooling is applied to the feature $\mathbf{F}_c$, which is formed by concatenating the multimodal feature $\mathbf{F}_i$ along the channel dimension ($\mathbf{F}_c=\mathrm{Concat}(\mathbf{F}_1,\mathbf{F}_2,\ldots,\mathbf{F}_8)$). Three 1D convolutions (kernel size = 3) are then used to capture local channel relationships, producing query, key, and value. Multi-head self-attention is subsequently employed to capture global interactions across channels. The global features are passed through a Sigmoid function to constrain them within the range $(0,1)$, and these values are used as weights to module $\mathbf{F}_c$, resulting in task-specific features $\mathbf{F}_{\mathrm{sp}_j}$. This design leverages the fact that different channels in $\mathbf{F}_c$ correspond to distinct modality features, such as those from MARNet and 3D-CNN, which are selectively emphasized depending on the target task (e.g., facial and body images are crucial for driver behavior and emotion recognition, as shown in the ablation study in Section \ref{Ablation studies on multimodal data}). The channel interaction, guided by global-local attention, dynamically adjusts the importance of each modality for different tasks, alleviating interference from irrelevant modalities. The computation process of the task-specific branch $\mathbf{T}_{\mathrm{sp}}$ is as follows:
\begin{equation}
\label{sp}
        \mathbf{F}_{\mathrm{sp}_j} = \mathbf{T}_{\mathrm{sp}}(\mathbf{F}_c) = \mathbf{F}_c \otimes \sigma(\mathrm{{MHA}}_j(\mathrm{{C1D}}_j(\mathrm{GAP}(\mathbf{F}_c)), n)) ,
\end{equation}
where $\mathrm{MHA}_j(a,b)$ denotes the multi-head self-attention for task $j$, $\mathrm{{C1D}}_j(\cdot)$ denotes the 1D convolution, $n$ is the number of heads, and $\otimes$ indicates element-wise multiplication. 

To enhance task synergy while extracting task-specific features, we designed the task-shared branch, as shown in Fig.\ref{SME} (b). To improve extraction efficiency, we first categorize multimodal features based on their source and structure. Features related to the traffic context (i.e., images from the left, right, and front views) are concatenated into $\mathbf{F}_{\mathrm{sc}}$, driver-related features (i.e., images from the inside view, driver's face, and body) form $\mathbf{F}_{\mathrm{dr}}$, and joint features (i.e., posture and gesture) form $\mathbf{F}_{\mathrm{jo}}$. To address scale differences in multimodal data, the task-shared branch combines $\mathbf{F}_{\mathrm{sc}}$ and $\mathbf{F}_{\mathrm{dr}}$, then applies convolutions with kernel sizes of 1 and 3 to capture features at different scales, producing a multi-scale feature map. The computation process $f(\cdot)$ is shown as follows:
\begin{equation}
f(\mathbf{F}_{\mathrm{dr}},\mathbf{F}_{\mathrm{sc}}) = \mathrm{C2D}_{1*1}(\mathbf{F}_{\mathrm{dr}} + \mathbf{F}_{\mathrm{sc}}) + \mathrm{C2D}_{3*3}(\mathbf{F}_{\mathrm{dr}} + \mathbf{F}_{\mathrm{sc}}),
\end{equation}
Where $\mathrm{C2D}_{1*1}$ and $\mathrm{C2D}_{3*3}$ represent 2D convolution operations with kernel sizes of 1 and 3, respectively. The task-specific branch then dynamically integrates this feature map and passes it through a Sigmoid function to generate weights. These weights are used to merge $\mathbf{F}_{\mathrm{sc}}$ and $\mathbf{F}_{\mathrm{dr}}$ adaptively, resulting in the preliminary shared features $\mathbf{F}_{\mathrm{ps}}$. The computation process of the task-shared branch $\mathbf{T}_{\mathrm{sh}}(\cdot)$ is shown as follows:
\begin{align}
    \mathbf{F}_{\mathrm{ps}} = \mathbf{T}_{\mathrm{sh}}(\mathbf{F}_{\mathrm{dr}}, \mathbf{F}_{\mathrm{sc}}) &= \mathbf{F}_{\mathrm{sc}} \times \sigma(\mathbf{T}_{\mathrm{sp}}(f(\mathbf{F}_{\mathrm{dr}} , \mathbf{F}_{\mathrm{sc}})))  \\ 
    & + \mathbf{F}_{\mathrm{dr}} \times (1 - \sigma(\mathbf{T}_{\mathrm{sp}}(f(\mathbf{F}_{\mathrm{dr}} , \mathbf{F}_{\mathrm{sc}}))))\notag .
\end{align}

Similarly, $\mathbf{F}_{\mathrm{ps}}$ and $\mathbf{F}_{\mathrm{jo}}$ undergo feature extraction by the task-shared branch using the same operations, resulting in the final task-shared features $\mathbf{F}_{\mathrm{sh}}$. The calculation is by:
\begin{equation}
    \mathbf{F}_{\mathrm{sh}} = \mathbf{T}_{\mathrm{sh}}(\mathbf{F}_{\mathrm{jo}}, \mathbf{F}_{\mathrm{ps}}).
\end{equation}

Finally, the task-specific features $\mathbf{F}_{\mathrm{sp}_j}$ and the task-shared features $\mathbf{F}_{\mathrm{sh}}$ are dynamically fused using Eq. (\ref{Oj}) to yield the recognition result $O_j$ for each task.

\begin{table*}[t]
\setlength{\tabcolsep}{4pt}
\centering
\renewcommand{\arraystretch}{1.2}
\vspace{-1em}
\resizebox{\linewidth}{!}{%
\begin{tabular}{c|cccccl|c|c|c|c|c}
\toprule
 & \multicolumn{6}{c|}{\textbf{Backbone}} & \textbf{DER} & \textbf{DBR} & \textbf{TCR} & \textbf{VBR} &  \\ \cline{2-11}
\multirow{-2}{*}{\centering \textbf{Pattern}} & \textbf{Face} & \textbf{Body} & \textbf{Gesture} & \textbf{Posture} & \multicolumn{2}{c|}{\textbf{Scene}} & \textbf{Acc} & \textbf{Acc} & \textbf{Acc} & \textbf{Acc} & \multirow{-2}{*}{\centering \textbf{mAcc}}\\ 
\midrule

\multirow{10}{*}{2D}  & Res18~\cite{he2016deep}         & Res34~\cite{he2016deep}            & MLP+SE  & MLP+SE  & \multicolumn{2}{c|}{PP-Res18~\cite{zhou2017places}}         & 69.05          & 63.87          & 88.01          & 78.16         & 74.77                 \\
 & Res18~\cite{he2016deep}            & Res34~\cite{he2016deep}            & MLP+SE  & MLP+SE  & \multicolumn{2}{c|}{Res34~\cite{he2016deep}}            & 71.26 & 65.35& 83.74          & 77.12          & 74.37               \\
& Res34~\cite{he2016deep}            & Res50~\cite{he2016deep}            & MLP+SE  & MLP+SE  & \multicolumn{2}{c|}{Res50~\cite{he2016deep}}            & 69.68          & 59.77         & 80.13          & 71.26         & 70.21                \\
 & VGG13~\cite{simonyan2014very}            & VGG16~\cite{simonyan2014very}            & MLP+SE  & MLP+SE  & \multicolumn{2}{c|}{VGG16~\cite{simonyan2014very}}            & 70.72          & 63.65          & 82.77          & 77.94         & 73.77                  \\
 & VGG16~\cite{simonyan2014very}            & VGG19~\cite{simonyan2014very}            & MLP+SE  & MLP+SE  & \multicolumn{2}{c|}{VGG19~\cite{simonyan2014very}}            & 69.31          & 62.34          & 83.58          &75.13          & 72.59       \\
 & CPVT~\cite{chu2023conditional}           & CPVT~\cite{chu2023conditional}              & ST-GCN  & ST-GCN  & \multicolumn{2}{c|}{CPVT~\cite{chu2023conditional}}              & 69.01          & 67.35          & 91.44          & 79.57 & 76.84  \\
 & CMT~\cite{Guo_2022_CVPR}           & CMT~\cite{Guo_2022_CVPR}              & ST-GCN  & ST-GCN  & \multicolumn{2}{c|}{CMT~\cite{Guo_2022_CVPR}}              & 68.75          & 68.75          & \underline{93.75}          & \underline{81.38} & 78.16  \\
 & GroupMixFormer~\cite{ge2023advancing}           & GroupMixFormer~\cite{ge2023advancing}              & ST-GCN  & ST-GCN  & \multicolumn{2}{c|}{GroupMixFormer~\cite{ge2023advancing}}              & 66.29          & 67.54          & 92.12          & 77.63 & 75.90  
 \\  & AbSViT~\cite{liu2024vision}           & AbSViT~\cite{liu2024vision}              & ST-GCN  & ST-GCN  & \multicolumn{2}{c|}{AbSViT~\cite{liu2024vision}}              & 69.15          & 67.84          & 92.07          & 80.82 & 77.47
 \\  & GLMDriveNet~\cite{liu2024glmdrivenet}             & GLMDriveNet~\cite{liu2024glmdrivenet}              & ST-GCN  & ST-GCN  & \multicolumn{2}{c|}{GLMDriveNet~\cite{liu2024glmdrivenet}}              & 71.38          & 66.57          & 90.23          & 77.19          & 76.34 
 \\ \midrule
\multirow{5}{*}{\begin{tabular}[c]{@{}c@{}}2D +\\ Timing\end{tabular}} & Res18~\cite{he2016deep}+TransE     & Res34~\cite{he2016deep}+TransE~\cite{vaswani2017attention}     & MLP+TE  & MLP+TE  & \multicolumn{2}{c|}{PP-Res18+TransE~\cite{vaswani2017attention}}  & 70.83          & 67.32          & 90.54          & 79.97         & 77.17          \\
& Res18~\cite{he2016deep}+TransE~\cite{vaswani2017attention}     & Res34~\cite{he2016deep}+TransE~\cite{vaswani2017attention}     & MLP+TE  & MLP+TE  & \multicolumn{2}{c|}{Res34~\cite{he2016deep}+TransE}       & 72.65            & 67.08    & 86.63          & 78.46         & 76.21                  \\
 & Res34~\cite{he2016deep}+TransE~\cite{vaswani2017attention}     & Res50~\cite{he2016deep}+TransE~\cite{vaswani2017attention}     & MLP+TE  & MLP+TE  & \multicolumn{2}{c|}{Res50~\cite{he2016deep}+TransE~\cite{vaswani2017attention}}     & 70.24          & 65.65          & 82.57          & 77.29          & 73.94                 \\
 & VGG13~\cite{simonyan2014very}+TransE~\cite{vaswani2017attention}     & VGG16~\cite{simonyan2014very}+TransE~\cite{vaswani2017attention}     & MLP+TE  & MLP+TE  & \multicolumn{2}{c|}{VGG16~\cite{simonyan2014very}+TransE~\cite{vaswani2017attention}}     & 71.12          & 67.15         & 85.13          & 78.58         & 75.50                \\
 & VGG16~\cite{simonyan2014very}+TransE~\cite{vaswani2017attention}     & VGG19~\cite{simonyan2014very}+TransE~\cite{vaswani2017attention}     & MLP+TE  & MLP+TE  & \multicolumn{2}{c|}{VGG19~\cite{simonyan2014very}+TransE~\cite{vaswani2017attention}}     & 69.46          & 65.48         & 85.74          & 77.91         & 74.65                       \\ \midrule
\multirow{11}{*}{3D}  & MobileNet-V1~\cite{howard2017mobilenets}  & MobileNet-V1~\cite{howard2017mobilenets}  & ST-GCN   & ST-GCN   & \multicolumn{2}{c|}{MobileNet-V1~\cite{howard2017mobilenets}}  & 72.23          & 64.20          & 88.34          &77.83          & 75.65              \\
& MobileNet-V2~\cite{sandler2018mobilenetv2}  & MobileNet-V2~\cite{sandler2018mobilenetv2}  & ST-GCN   & ST-GCN   & \multicolumn{2}{c|}{MobileNet-V2~\cite{sandler2018mobilenetv2}}  & 68.47          & 61.74         & 86.54          & 78.66        & 73.85               \\
 & ShuffleNet-V1~\cite{zhang2018shufflenet} & ShuffleNet-V1~\cite{zhang2018shufflenet} & ST-GCN   & ST-GCN   & \multicolumn{2}{c|}{ShuffleNet-V1~\cite{zhang2018shufflenet}} & 72.41          & \underline{68.97}         & 90.64          &80.79         & \underline{78.20}              \\
 & ShuffleNet-V2~\cite{ma2018shufflenet} & ShuffleNet-V2~\cite{ma2018shufflenet} & ST-GCN   & ST-GCN   & \multicolumn{2}{c|}{ShuffleNet-V2~\cite{ma2018shufflenet}} & 70.94          & 64.04          & 89.33          & 78.98         & 75.82               \\
& 3D-Res18~\cite{hara2018can}         & 3D-Res34~\cite{hara2018can}         & ST-GCN   & ST-GCN   & \multicolumn{2}{c|}{3D-Res34~\cite{hara2018can}}         & 70.11          & 66.52          & 88.51          & 81.21          & 76.59                 \\
  & 3D-Res34~\cite{hara2018can}         & 3D-Res50~\cite{hara2018can}         & ST-GCN   & ST-GCN   & \multicolumn{2}{c|}{3D-Res50~\cite{hara2018can}}         & 69.13          & 63.05          & 87.82          & 79.31          & 74.83                \\
& C3D~\cite{tran2015learning}              & C3D~\cite{tran2015learning}              & ST-GCN   & ST-GCN   & \multicolumn{2}{c|}{C3D~\cite{tran2015learning}}              & 63.05          & 63.95         & 85.41          & 77.01        & 72.36                 \\
 & I3D~\cite{carreira2017quo}              & I3D~\cite{carreira2017quo}               & ST-GCN   & ST-GCN   & \multicolumn{2}{c|}{I3D~\cite{carreira2017quo} }              & 70.94          &66.17         & 87.68         & 79.81         & 76.15                \\
  & SlowFast~\cite{feichtenhofer2019slowfast}         & SlowFast~\cite{feichtenhofer2019slowfast}         & ST-GCN   & ST-GCN   & \multicolumn{2}{c|}{SlowFast~\cite{feichtenhofer2019slowfast}}         & 72.38          & 61.58         & 86.86          & 78.33          & 74.79       \\

 & TimeSFormer~\cite{bertasius2021space}      & TimeSFormer~\cite{bertasius2021space}      & ST-GCN   & ST-GCN   & \multicolumn{2}{c|}{TimeSFormer~\cite{bertasius2021space}}      & \underline{74.87}  & 65.18 & 92.12 & 78.81 & 77.75                 \\
   & Video Swin Transformer~\cite{liu2022video}         & Video Swin Transformer~\cite{liu2022video}        & 3DCNN    & 3DCNN    & \multicolumn{2}{c|}{Video Swin Transformer~\cite{liu2022video}}          & 73.44          & 65.63          & \underline{93.75}          & 75.00          & 76.96                \\ \midrule
 \rowcolor{gray!15}
\multirow{1}{*}{\textbf{Ours}} & \textbf{MARNet}         & \textbf{MARNet}             & \textbf{3DCNN}    & \textbf{3DCNN}    & \multicolumn{2}{c|}{\textbf{MARNet}}          & \textbf{76.67          }& \textbf{73.61}          & \textbf{93.91}          & \textbf{85.00}          & \textbf{82.30}    \\ \bottomrule
\end{tabular}
}
\caption{Comparison results of baselines on the AIDE dataset for all four tasks. DER is Driver Emotion Recognition, DBR is Driver Behavior Recognition, TCR is Traffic Context Recognition, VBR is Vehicle Behavior Recognition. The best results is highlighted in \textbf{bold}, while the second-best results are \underline{underlined}. The row highlighted in \cellcolor{gray!15} \raisebox{0.8ex}{\fcolorbox{gray!15}{gray!15}{\hspace{0.4cm}}} indicates our proposed method. Scene denotes multi-view images (i.e., front-view, right-view, left-view, inside-view). \textbf{Res}: ResNet~\cite{he2016deep}, \textbf{MLP}: multi-layer perception, \textbf{SE}: spatial embedding, \textbf{TE}: temporal embedding, \textbf{TransE}: transformer encoder~\cite{vaswani2017attention}.}
\label{table1}
\vspace{-0.4cm}
\end{table*}

\subsection{Loss Function}

In MMTL-UniAD, we propose a novel loss function that integrates the individual losses of each task to optimize overall model performance. Specifically, the total loss $L_{\mathrm{total}}$ is calculated as follows:
\begin{equation}
L_{\mathrm{total}} = \sum_{j=1}^{m} \mathrm{CrossEntropy}(O_j, \mathrm{label}_j),
\end{equation}
where $O_j$ represents the recognition results for task $j$, $\mathrm{label}_j$  denotes the corresponding ground truth label, $\mathrm{CrossEntropy}$ is the cross-entropy loss function. The number of tasks, $m$, is set to 4, corresponding to the four recognition tasks implemented: driver emotion recognition, driver behavior recognition, traffic context recognition, and vehicle behavior recognition.

\section{Experiments}
We conducted extensive experiments using the open-source AIDE dataset \cite{yang2023aide} to evaluate the effectiveness of our proposed MMTL-UniAD for multi-task learning. This section introduces the AIDE dataset, data preprocessing, and evaluation metrics, followed by a performance comparison of MMTL-UniAD across four tasks relative to existing algorithms. Finally, we present and analyze the results of our ablation experiment.

\subsection{Dataset}

The AIDE dataset consists of 2,898 samples with multi-view, multimodal, and multi-task features. Each sample includes multi-view video data (i.e., front, right, left, inside views) and multimodal data related to the driver (i.e., images of the driver's face and body, and joint data representing driver posture and gestures). The dataset is split into training (65\%), validation(15\%), and test (20\%) sets. Annotations cover four tasks: driver emotion recognition, driver behavior recognition, traffic context recognition, and vehicle behavior recognition.

\subsection{Data Preprocessing}

For data preprocessing, we used bounding box coordinates to crop inside-view images, extracting refined images of the driver’s face and body. We then synchronized multi-view video data and joint data at 16 frames per second to ensure temporal alignment. Each model receives a sequence of 16 consecutive frames along with corresponding joint data for each frame, preserving temporal information during feature extraction. Additionally, data augmentation was applied to all images with a 50\% probability of random horizontal and vertical flipping.

\subsection{Evaluation Metrics}

To assess the model performance, we adopted accuracy (\%) as the primary metric. Furthermore, following \cite{fan2022mlfnet,shi2023bssnet}, we introduced a new evaluation metric $\mathrm{mAcc}$ (\%) for multi-task learning, defined as:
\begin{equation}
    \mathrm{mAcc} = \frac{1}{m} \sum_{j=1}^{m} \mathrm{Acc}_j,
\end{equation}
where $m$ is the number of tasks in the multi-task learning, and $\mathrm{Acc}_j$ is the accuracy achieved by the model on task $j$. We use mAcc to provide a comprehensive evaluation across all tasks, ensuring a balanced assessment of model performance rather than focusing on individual tasks.

\subsection{Implement Details}

All experiments were conducted on an NVIDIA A40 GPU with a batch size of 24 for both training and testing. The stochastic gradient descent (SGD) optimizer was applied with a momentum of 0.9 and weight decay of 0.0001. The initial learning rate was set to 0.1, and the number of epochs was 125. For region attention, the $t$ was set to 7, indicating a region size of $7 \times 7$, and $k$ was set to 4, selecting the top 4 regions with the highest similarity to the target region.

\subsection{Comparison with the State-of-the-Art}

Table \ref{table1} presents the multi-task evaluation results of our proposed MMTL-UniAD, compared to the state-of-the-art methods. Based on the feature extraction dimensions of multi-view sequential images, we categorize the comparison methods into three patterns, following \cite{yang2023aide}: \textbf{2D}(using 2D models, e.g., VGG \cite{simonyan2014very}, ResNet\cite{he2016deep}, and CMT \cite{Guo_2022_CVPR}), \textbf{2D+Timing} (combining 2D models with sequence models \cite{vaswani2017attention}), and \textbf{3D} (using 3D models, e.g., Video Swin Transformer \cite{liu2022video}, and 3D Implementations of MobileNet \cite{howard2017mobilenets} and ShuffleNet \cite{zhang2018shufflenet}). 

Our MMTL-UniAD outperforms all algorithms in these three patterns, improving the mAcc metric by 4.10\%-12.09\%, and achieving the best results across all four tasks. Notably, it significantly improves recognition accuracy for driver behavior (by 4.64\%) and vehicle behavior (by 3.62\%), which demonstrates the superiority of our algorithm in multi-task learning.

The outstanding performance of MMTL-UniAD can be attributed to the design of its MARNet and dual-branch multimodal embedding. These components leverage multi-attention mechanisms and a dual-branch structure for task-shared and task-specific feature extraction. This design enhances cross-task knowledge transfer while mitigating negative transfer caused by inter-task differences and irrelevant features. In contrast, other algorithms, although employing various feature extraction backbones for multi-task recognition, are not optimized to address multi-task learning challenges, such as negative transfer. As a result, they suffer from significant negative transfer effects between tasks, limiting improvements in recognition performance.

\begin{table}[t]
\def\arraystretch{1.25}
\resizebox{\linewidth}{!}{%
\begin{tabular}{c|cccc|cccc}
\toprule
& \multicolumn{4}{c|}{\textbf{Task}} & \textbf{DER} & \textbf{DBR} & \textbf{TCR} & \textbf{VBR} \\ \cline{2-9}
\multirow{-2}{*}{\centering \textbf{Method}} & 
\multicolumn{2}{c}{\textbf{Driver States}} & 
\multicolumn{2}{c|}{\textbf{Traffic Context}} & 
\textbf{Acc} & \textbf{Acc} & \textbf{Acc} & \textbf{Acc} \\ 
\midrule

\multirow{2}{*}{Contrast}
& \multicolumn{2}{c}{w/} &\multicolumn{2}{c|}{w/o} &  72.22 & 69.35 & - & - \\

& \multicolumn{2}{c}{w/o}& \multicolumn{2}{c|}{w/} & - & - & 90.41 & 80.63 \\ \midrule

\rowcolor{gray!15}\multicolumn{1}{c|}{\textbf{Ours}} & \multicolumn{2}{c}{w/} & \multicolumn{2}{c|}{w/}  & \textbf{76.67} & \textbf{73.61} & \textbf{93.91} & \textbf{85.00} \\ \bottomrule

\end{tabular}
}
\caption{Results of Multi-task Ablation Experiments for driver states and traffic context. "w/" indicates the use of the corresponding component or method, while "w/o" means the component or method was not used. Driver states include DER and DBR, while the traffic context includes TCR and VBR.}
\vspace{-0.4cm}
\label{table15}
\end{table}

\subsection{Ablation Experiment}
We conduct a series of ablation experiments to assess the effectiveness and individual contributions of multi-task learning, multimodal data, and key components, including MARNet and the dual-branch multimodal embedding.

\subsubsection{Ablation Studies on Multi-task Learning}
To verify the necessity of multi-task learning, we perform two sets of ablation experiments. 

The first set evaluates the impact of including both driver states and traffic context tasks. Results in Table \ref{table15} show that when only the driver states-related tasks are jointly trained, without considering traffic context tasks, the accuracy of driver emotion recognition and driver behavior recognition decreases by 4.26\%-4.45\%. Similarly, when only traffic context-related tasks are trained, excluding driver states, the accuracy of traffic context and vehicle behavior recognition drops by 3.50\%-4.37\%. These results demonstrate that features learned from the driver states tasks benefit the traffic context recognition tasks, and vice versa, highlighting the advantages of joint learning for improving model accuracy and generalization.

The second set of experiments examines the interactions between different tasks. The first part trains the model on a single task, while the second part removes one task, leaving the other three. Table \ref{table3} shows that training on a single task results in a 3.98\%-6.13\% drop in performance across all tasks. Moreover, removing any one of the four tasks also reduce the accuracy of the remaining tasks. These findings underscore the positive impact of jointly learning all four tasks, further validating the interrelated nature of these tasks and the effectiveness of multi-task learning.

\begin{table}[t]
\centering
\renewcommand{\arraystretch}{1.2} 
\resizebox{\linewidth}{!}{%
\begin{tabular}{>{\centering\arraybackslash}m{1.2cm}|>{\centering\arraybackslash}m{1.2cm}|>{\centering\arraybackslash}m{4cm}|>{\centering\arraybackslash}m{0.6cm} >{\centering\arraybackslash}m{0.6cm} >{\centering\arraybackslash}m{0.6cm} >{\centering\arraybackslash}m{0.6cm}}
\toprule
& & & \textbf{DER} & \textbf{DBR} & \textbf{TCR} & \textbf{VBR} \\ \cline{4-7}
\multirow{-2}{*}{\centering \textbf{Method}} & \multirow{-2}{*}{\centering \textbf{Config}} & \multirow{-2}{*}{\centering \textbf{Task}} & \textbf{Acc} & \textbf{Acc} & \textbf{Acc} & \textbf{Acc} \\ 
\midrule

\multirow{8}{*}{Contrast}
& w/o & DBR \ \& \ TCR \ \& \ VBR & 70.56 & - & - & - \\

& w/o & DER \ \& \ TCR \ \& \ VBR & - & 68.12 & - & - \\

& w/o & DER \ \& \ DBR  \ \& \ VBR & - & - & 89.93 & - \\

& w/o & DER \ \& \ DBR \ \& \ TCR & - & - & - & 78.87 \\

& w/ & DBR \ \& \ TCR \ \& \ VBR & - & 64.19 & 92.36 & 83.71 \\

& w/ & DER \ \& \ TCR \ \& \ VBR & 73.01 & - & 91.73 & 84.56 \\ 

& w/ & DER \ \& \ DBR \ \& \ VBR & 75.32 & 70.83 & - & 79.85 \\

& w/ & DER \ \& \ DBR \ \& \ TCR & 74.62 & 67.24 & 89.03 & - \\ \midrule
\rowcolor{gray!15} 
\multirow{1}{*}{\textbf{Ours}} & w/ & Full  Tasks
& \textbf{76.67} & \textbf{73.61} & \textbf{93.91} & \textbf{85.00} \\ \bottomrule

\end{tabular}
}
\caption{Results of Multi-task Ablation Experiments.}
\label{table3}
\vspace{-0.5cm}
\end{table}

\subsubsection{Ablation Studies on MARNet and Dual-Branch Multimodal Embedding}
We conducted ablation experiments to assess the individual contributions of the key components, MARNet and dual-branch multimodal embedding, within MMTL-UniAD. Specifically, we replaced MARNet with a basic VGG network and replaced the dual-branch multimodal embedding with a simple concatenation operation. The results in Table \ref{table4} show that models without MARNet or the dual-branch multimodal embedding exhibit a significant performance drop of 5.34\%-12.05\% in the mAcc metric, with accuracy across all four tasks decreasing accordingly. 

This degradation is attributed to the failure of the alternative components to effectively capture key features during extraction, resulting in the inclusion of task-unrelated information in the multimodal features. This not only weakens the task synergy but also exacerbates negative transfer due to task conflicts. These findings demonstrate that the inclusion of MARNet and the dual-branch multimodal embedding is essential for optimal performance, as they facilitate better feature extraction and task-specific learning, ultimately enhancing cross-task synergy and the overall effectiveness of multi-task learning.

\subsubsection{Ablation Studies on Multimodal Data}
\label{Ablation studies on multimodal data}
To further investigate the contribution of each modality, we performed ablation experiments using different combinations of input modalities. The input data were divided into three groups: (i) facial and body sequential images of the driver, (ii) gesture and body joint data, and (iii) multi-view sequential images (i.e., right, left, front, inside views). Each group was used individually to train the model, and the results are presented in Table \ref{table2}.

The results show a notable decrease in both mAcc and accuracy across all tasks when only one modality was used, highlighting the importance of multimodal data for these recognition tasks. This fully demonstrates the critical role of using multimodal data in improving the performance of multi-task learning models and the need for multi-task learning architectures specifically designed to integrate multimodal inputs effectively.

\begin{table}[t]
\setlength{\tabcolsep}{7pt}
\centering
\resizebox{\linewidth}{!}{%
\begin{tabular}{c|cc|cccc|c}
\toprule                                              
& & & \textbf{DER} & \textbf{DBR} & \textbf{TCR} & \textbf{VBR} &  \\ \cline{4-7}
\multirow{-2}{*}{\centering \textbf{Method}} & \multirow{-2}{*}{\centering \textbf{MARNet}} & \multirow{-2}{*}{\centering \textbf{DBME}} & \centering \textbf{Acc} & \centering \textbf{Acc} & \centering \textbf{Acc} & \centering \textbf{Acc} & \multirow{-2}{*}{\centering \textbf{mAcc}}\\ 
\midrule

\multirow{3}{*}{Contrast} & w/o  &  w/o  &  62.91     & 60.73     & 82.84      & 74.33      & 70.25     \\
& w/   &  w/o  & 71.66      & 67.12     & 91.50      & 77.56      & 76.96     \\
& w/o  &  w/   & 70.35     & 68.45      & 89.22      & 79.12   & 76.79 \\ \midrule
\rowcolor{gray!15} \multirow{1}{*}{\textbf{Ours}} &
w/   &  w/   & \textbf{76.67}& \textbf{73.61}          & \textbf{93.91}          & \textbf{85.00}          & \textbf{82.30}     \\

\bottomrule
\end{tabular}
}
\caption{Ablation experiment results of MARNet and Dual-Branch Multimodal Embedding. Among them, DBME represents Dual-Branch Multimodal Embedding}
\label{table4}
\end{table}

\begin{table}[t]
\setlength{\tabcolsep}{7pt}
\renewcommand{\arraystretch}{1.2}
\centering
\resizebox{\linewidth}{!}{%
\begin{tabular}{c|cccl|cccc|c}
\toprule
& \multicolumn{4}{c|}{\textbf{Multimodal Data}} & \textbf{DER} & \textbf{DBR} & \textbf{TCR} & \textbf{VBR} &  \\ \cline{2-9}
\multirow{-2}{*}{\centering \textbf{Method}} & \multicolumn{2}{c}{\centering \textbf{Face+Body}} & \centering \textbf{G+P} & \textbf{Scene} & \textbf{Acc} & \textbf{Acc} & \textbf{Acc} & \textbf{Acc} & \multirow{-2}{*}{\centering \textbf{mAcc}}\\ 
\midrule

\multirow{3}{*}{Contrast}

&\multicolumn{2}{c}{}  &      &\multicolumn{1}{c|}{\CheckmarkBold} & 67.81   & 64.32    & 92.87    & 82.66   & 76.91    \\

&\multicolumn{2}{c}{}  &\multicolumn{1}{c}{\CheckmarkBold}        &\multicolumn{1}{c|}{}     & 68.22          & 54.13          & 63.11          & 37.29  & 55.69        \\

&\multicolumn{2}{c}{\CheckmarkBold}  &        &\multicolumn{1}{c|}{}     & 72.91          & 71.77          & 89.16          & 78.69  & 78.13        \\

\midrule
\rowcolor{gray!15} 
\multirow{1}{*}{\textbf{Ours}}                                   & \multicolumn{2}{c}{\CheckmarkBold}           & \multicolumn{1}{c}{\CheckmarkBold}  
& \multicolumn{1}{c|}{\CheckmarkBold}          & \textbf{76.67}& \textbf{73.61}          & \textbf{93.91}          & \textbf{85.00}          & \textbf{82.30}                            \\ 
 \bottomrule
\end{tabular}
}
\caption{Results of Multimodal Ablation  Experiments for Four Tasks. G+P represents gesture and posture joint data.}
\label{table2}
\vspace{-0.4cm}
\end{table}
\section{Conclusions}

This paper proposes MMTL-UniAD, a unified multi-modal multi-task learning framework. This framework leverages multi-modal data for simultaneous recognition of driver emotions, driver behavior, traffic context, and vehicle behavior. Central to this framework is the multi-axis region attention network and dual-branch multimodal embedding, which facilitate the effective extraction of both task-shared and task-specific features. These components not only enhance cross-task synergy but also mitigate the negative transfer, leading to superior performance across all four tasks on the open-source AIDE dataset. Ablation experiments further highlight that joint learning of driver states and traffic context-related tasks enables mutual feature sharing, which significantly improves task recognition accuracy. We anticipate that MMTL-UniAD along with its core components will serve as a robust baseline for future research in multimodal multi-task learning for Advanced Driver Assistance Systems (ADAS), advancing the development of more effective and adaptable systems in this domain.
{
    \small
    \bibliographystyle{ieeenat_fullname}
    \bibliography{main}
}

\end{document}